\documentclass[10pt,twocolumn,letterpaper]{article}

\usepackage{iccv}

\usepackage{times}
\usepackage{epsfig}
\usepackage{graphicx}
\usepackage{amsmath}
\usepackage{amssymb}
\usepackage{multirow}
\usepackage{subcaption}

\usepackage{comment} % added
\usepackage[accsupp]{axessibility}

\definecolor{iccvblue}{rgb}{0.21,0.49,0.74}
\usepackage[pagebackref,breaklinks,colorlinks,allcolors=iccvblue]{hyperref}

\def\paperID{} % *** Enter the Paper ID here
\def\confName{ICCV}
\def\confYear{2025}

\begin{document}

%%%%%%%%% TITLE
% \title{3D Occupancy Prediction Mdels Efficiently Pre-trained with 3D Occcupancy Dataset without Semantic Annotation}
% \title{Efficient Pre-training of 3D Occupancy Prediction Models Using Unannotated Occupancy Datasets}
% \title{Efficient Pre-training of 3D Occupancy Prediction Models Using Binary Occupancy Datasets}

\title{From Binary to Semantic: Utilizing Large-Scale Binary Occupancy Data \\ for 3D Semantic Occupancy Prediction}

% \title{Efficient Pre-training of 3D Occupancy Prediction Models with Binary Occupancy Datasets}

% \title{Pre-training of 3D Occupancy Prediction Models with Binary Occupancy Datasets}

% \title{Efficient Pre-training of 3D Occupancy Prediction Model Using Binary Occupancy Dataset}

% \title{Pre-training strategy of 3D Occupancy Prediction Models Using Binary Occupancy Datasets}

\author{Chihiro Noguchi$^*$ \qquad Takaki Yamamoto$^*$\\
InfoTech,\, Toyota Motor Corporation\\
{\tt\small \{chihiro\_noguchi\_aa, takaki\_yamamoto\}@mail.toyota.co.jp}
% \and
% Takaki Yamamoto \\
% Toyota Motor Corporation\\
}

\maketitle

\def\thefootnote{*}\footnotetext{Equal Contribution.}

%%%%%%%%% ABSTRACT
\begin{abstract}
Accurate perception of the surrounding environment is essential for safe autonomous driving. 3D occupancy prediction, which estimates detailed 3D structures of roads, buildings, and other objects, is particularly important for vision-centric autonomous driving systems that do not rely on LiDAR sensors. However, in 3D semantic occupancy prediction—where each voxel is assigned a semantic label—annotated LiDAR point clouds are required, making data acquisition costly. In contrast, large-scale binary occupancy data, which only indicate occupied or free space without semantic labels, can be collected at a lower cost. Despite their availability, the potential of leveraging such data remains unexplored. In this study, we investigate the utilization of large-scale binary occupancy data from two perspectives: (1) pre-training and (2) learning-based auto-labeling. We propose a novel binary occupancy-based framework that decomposes the prediction process into binary and semantic occupancy modules, enabling effective use of binary occupancy data. Our experimental results demonstrate that the proposed framework outperforms existing methods in both pre-training and auto-labeling tasks, highlighting its effectiveness in enhancing 3D semantic occupancy prediction. The code will be available at \url{https://github.com/ToyotaInfoTech/b2s-occupancy}

\end{abstract}

%%%%%%%%% BODY TEXT
\section{Introduction}

\begin{figure}[hbt]
\centering
\includegraphics[width=0.48\textwidth]{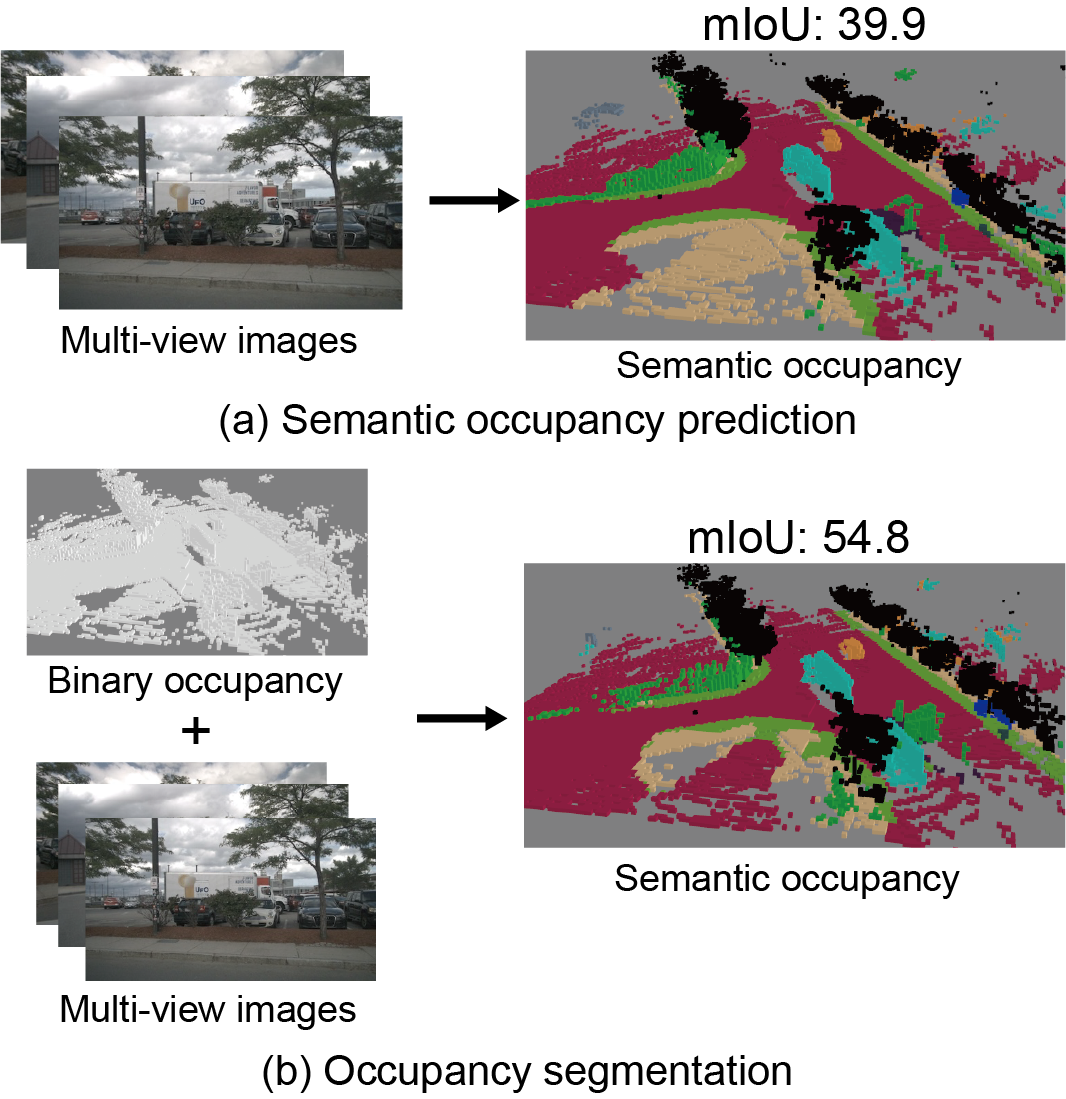}
\caption{Compared to (a) semantic occupancy prediction, (b) occupancy segmentation utilizes binary occupancy as an additional input along with multi-view images, leading to higher performance.}
\label{fig:fig_intro}
\end{figure}

A comprehensive understanding of the surrounding 3D scene is crucial for modern autonomous driving systems.  
While LiDAR sensors provide an effective solution for this purpose~\cite{zhou2018voxelnet, langPointPillarsFastEncoders2019a, yinCenterBased3DObject2021b, yanSparseSingleSweep2021}, their high cost remains a significant drawback. 
In contrast, vision-based methods have gained attention due to their affordability and recent technological advancements~\cite{li2024bevformer, cao2022monoscene, huang2023tri, huPlanningOrientedAutonomousDriving2023b,jiangVADVectorizedScene2023a}. 
This paper focuses on vision-based semantic occupancy prediction, which seeks to estimate both the occupancy state and the semantic labels of the surrounding 3D voxels relative to the ego-vehicle.

From a practical perspective, semantic occupancy prediction faces a significant challenge due to high annotation costs~\cite{wangOpenOccupancyLargeScale2023b, weiSurroundOccMultiCamera3D2023c, tianOcc3DLargeScale3D2023a}. 
Creating ground truth (GT) data for semantic occupancy prediction typically requires attaching semantic labels to 3D LiDAR points—a process that is both labor-intensive and costly, as it involves detailed and precise manual annotation.
However, if only binary occupancy information is needed (i.e., distinguishing between occupied and free space without semantic class labels), annotation costs can be substantially reduced. This is because semantic labels and fine-grained LiDAR annotation become unnecessary. For static scenes, binary occupancy data can be efficiently generated by aggregating LiDAR points over time, eliminating the need for point-wise semantic annotation. For dynamic objects, required data—such as 3D bounding boxes—can often be automatically created using auto-labeling systems for 3D object tracking and bounding box annotation~\cite{qi2021offboard,ma2023detzero,fan2023once}, which are commonly available in the development environments of end-to-end autonomous driving models~\cite{karnchanachariLearningbasedPlanningNuPlan2024c, caesarNuPlanClosedloopMLbased2022,sun2020scalability,waymochallenge2025}.
% Such tools are widely integrated into the development pipelines of end-to-end autonomous driving models~\cite{karnchanachariLearningbasedPlanningNuPlan2024c, caesarNuPlanClosedloopMLbased2022,sun2020scalability,waymochallenge2025}, further streamlining the annotation process. 
Although binary occupancy data can be collected at scale with relatively low effort, there has been limited research on how to effectively leverage them for semantic occupancy prediction.

% In summary, if a binary occupancy dataset is sufficient, there is no need for time-consuming semantic annotation of 3D bounding boxes or detailed point-wise labeling, significantly reducing both manual effort and cost. Conversely, semantic occupancy datasets require both 3D bounding boxes with semantic labels and point-wise semantic annotation, increasing the annotation burden, especially given the need for intensive quality control of auto-labeled data.

% We refer to this type of data, which represent only two states—occupied or free—as binary occupancy.
% Although binary occupancy data can be collected at scale with relatively low effort, there has been limited research on how to effectively leverage them for semantic occupancy prediction.

This study investigates how large-scale binary occupancy data can improve semantic occupancy prediction. As illustrated in Fig.~\ref{fig:fig_intro}, our core idea is that if the binary occupancy is known (Fig.~\ref{fig:fig_intro}(b)), the task is reduced to simply segmenting the occupied voxels. This is significantly simpler than the original task of simultaneously learning 3D geometry and semantics (Fig.~\ref{fig:fig_intro}(a)), and we expect it to lead to better performance. Our empirical results on the Occ3D dataset confirm this, showing that using ground truth (GT) binary occupancy data boosts the mean Intersection over Union (mIoU) from 39.9 to 54.8. However, because GT data is not available during online prediction, we propose two frameworks to approximate this process using large-scale binary occupancy data.
% In this study, we investigate how leveraging large-scale binary occupancy data can improve semantic occupancy prediction. 
% The fundamental concept of our approach is illustrated in Fig.~\ref{fig:fig_intro}. 
% Once the binary occupancy is provided (as shown in Fig.~\ref{fig:fig_intro}(b)), the task is reduced to segmenting the occupied voxels.
% Compared to the original task (Fig.~\ref{fig:fig_intro}(a)), which requires the simultaneous learning of both 3D structures and semantic labels, this task is simplified and expected to achieve higher performance. 
% In fact, our empirical experiments on the Occ3D dataset~\cite{tianOcc3DLargeScale3D2023a} demonstrate a significant increase in mean Intersection over Union (mIoU) from 39.9 to 54.8 when utilizing GT binary occupancy data in our proposed method. 
% Given that GT binary occupancy data is not available for online prediction, we introduce two frameworks that effectively realize the core concept through the utilization of large-scale binary occupancy data.

The first framework is a pre-training approach. We leverage large-scale binary occupancy data to improve the model's ability to predict geometric structures, thereby bringing the binary prediction performance closer to the GT level. Our experiments show that this pre-training is effective for the downstream task of semantic occupancy prediction. We also introduce a simple yet effective architecture optimized for this binary pre-training scheme.
% By enhancing the accuracy of binary occupancy prediction, the model can better approximate the ideal scenario envisioned in the core concept, thereby improving semantic occupancy prediction performance. However, our findings indicate that to fully exploit the benefits of pre-training, an architecture capable of effectively handling the sparsity of occupied voxels is crucial. To address this challenge, we propose a simple yet effective architecture specifically designed for binary occupancy pre-training, ensuring optimal performance in downstream semantic occupancy prediction tasks.

Our second framework involves learning-based auto-labeling with binary occupancy data, a crucial technique for creating large-scale datasets where manual annotation is cost-prohibitive. In this offboard context, which has no causality constraints and minimal inference speed restrictions, binary occupancy data can be used as an input. Our results show this scheme achieves prediction performance superior to existing online methods, including multi-modal approaches, enabling it to create more accurate pseudo-labels for auto-labeling systems. To our knowledge, this is the first study to introduce an offboard semantic occupancy prediction model.
% Second, we investigate learning-based auto-labeling utilizing binary occupancy data.  Auto-labeling plays a critical role in the creation of large-scale datasets, as manual annotation is prohibitively costly. In this context, we typically consider offboard settings, where there are no constraints on model causality and only minimal restrictions on inference speed. Under these conditions, binary occupancy data can be assumed to be available as input. Our experimental results demonstrate that the proposed scheme achieves superior prediction performance compared to existing online methods, including multi-modal approaches, and is highly effective in the distillation of online models. Furthermore, to the best of our knowledge, this study is the first to introduce an offboard semantic occupancy prediction model.

Our contributions are summarized as follows.
\begin{itemize}
    \item We propose a simple yet effective model architecture that utilizes large-scale binary occupancy data to enhance semantic occupancy prediction.
    \item We experimentally demonstrate that pre-training the proposed model with binary occupancy data outperforms prevalent pre-training methods.
    \item We further show that the proposed model performs effectively in an offboard setting. To the best of our knowledge, this study is the first to introduce an offboard semantic occupancy prediction model.
\end{itemize}

\section{Related Work}

\noindent
{\bf Vision-based 3D perception.} Bird’s-Eye-View (BEV) representation is a fundamental technology for vision-based 3D perception~\cite{li2024bevformer}. 
Two primary approaches are commonly used to obtain a BEV representation from image features. The first approach, Lift-Splat-Shoot (LSS~\cite{philionLiftSplatShoot2020b}), leverages depth distribution to project pixel-level image features into BEV space using camera calibration parameters. The second approach employs deformable attention modules, enabling BEV queries, defined in BEV space, to interact effectively with the corresponding image features~\cite{li2024bevformer}.

\noindent
{\bf 3D semantic occupancy prediction.} 
The goal is to assign semantic labels to each voxel in 3D space. A naive solution computes feature vectors for all voxels, but this is memory-intensive. To improve efficiency, many works explore alternative scene representations. Some use 2D planes—e.g., BEVFormer~\cite{li2024bevformer} (BEV) and TPVFormer~\cite{huang2023tri} (tri-perspective views). Others adopt coarse-to-fine strategies~\cite{tong2023scene,wei2023surroundocc,wang2024panoocc} to avoid full 3D grids. COTR~\cite{maCOTRCompactOccupancy2024a} proposes a compact 3D encoding. Sparse methods also gain traction, focusing on eliminating empty space~\cite{lu2023octreeocc,liu2025fully,tang2024sparseocc}. OSP~\cite{shi2024occupancy} and OPUS~\cite{wang2024opus} represent scenes as point sets with refinement, while GaussianFormer~\cite{huang2024gaussianformer} uses 3D Gaussians for efficient, flexible encoding. On the architectural side, OccFormer~\cite{zhang2023occformer} introduces a dual-path transformer inspired by Mask2Former~\cite{cheng2022masked}, and FB-OCC~\cite{li2023fb} combines LSS and BEV-based views. VoxFormer~\cite{li2023voxformer}, most similar to our approach, also follows a two-stage design but focuses on binary occupancy data rather than semantic granularity.

\noindent
{\bf Occupancy pre-training.} 
Pre-training is an important strategy in 3D recognition tasks. In the context of autonomous driving, it is common to consider problem settings where LiDAR points or high-precision 3D bounding boxes are available, and pre-training using these data is widely explored~\cite{liang2021exploring,yin2022proposalcontrast,yuan2023ad,boulch2023also,min2023occupancy}. Occupancy-based pre-training methods have also been proposed, with BEV segmentation~\cite{sirko2024occfeat} and planning~\cite{min2024driveworld} as downstream tasks. Additionally, UniScene~\cite{min2024multi} suggests a fine-tuning approach using a simple head exchange, although validation using large-scale occupancy data has not been conducted. This paper proposes an approach specifically focused on semantic occupancy prediction with large-scale occupancy pre-training.

\begin{figure*}[hbt]
\centering
\includegraphics[width=0.99\textwidth]{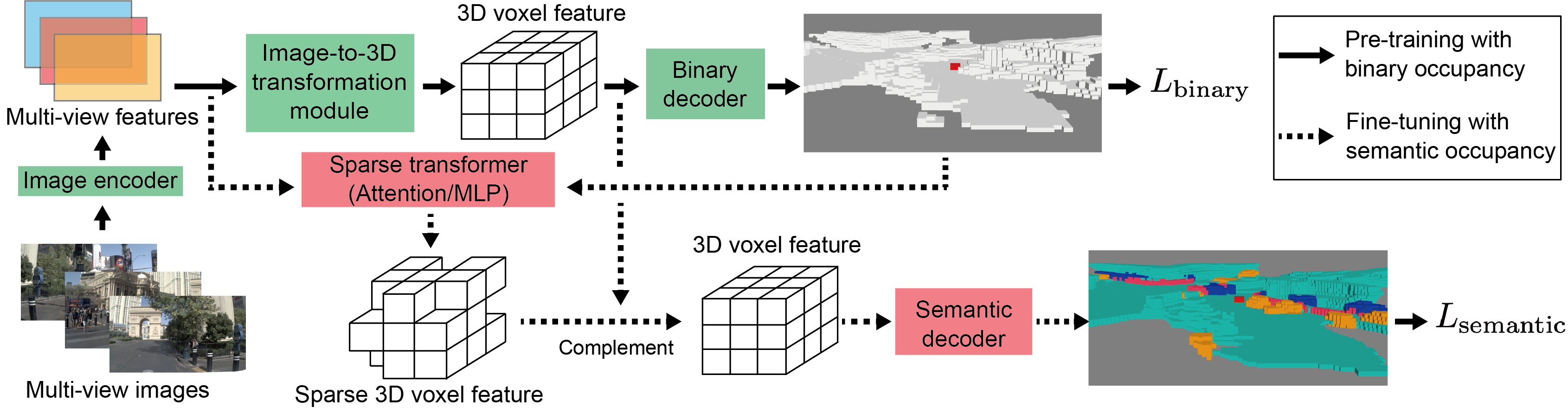}
\caption{Overview of the model architecture. Multi-view features extracted from an image encoder are processed through an image-3D transformation module to obtain a 3D voxel feature representation. Pre-training is performed up to the binary decoder (depicted with solid lines) using $L_{\rm{binary}}$, while fine-tuning is applied to the entire model with both $L_{\rm{binary}}$ and $L_{\rm{semantic}}$. For learning-based auto-labeling, GT binary occupancy is used in place of the binary decoder to identify occupied regions. Red voxels at the center of occupancy images represent the ego car. }
\label{fig:fig_concept}
\end{figure*}

\noindent
% {\bf Pretraining:} 
{\bf Multi-modal 3D semantic occupancy prediction.} 
While camera images provide rich semantic information, they lack precise geometric details. Consequently, leveraging the complementary strengths of LiDAR has become a widely adopted approach for enhancing 3D perception. Although this strategy has been extensively explored in the context of 3D object detection~\cite{chen2017multi,li2022deepfusion,chen2023bevdistill, liu2023bevfusion,li2024gafusion}, it is also highly effective for semantic occupancy prediction. 
M-CONet~\cite{wangOpenOccupancyLargeScale2023b} employs a coarse-to-fine strategy to enhance computational efficiency, while Co-Occ~\cite{wangOpenOccupancyLargeScale2023b} introduces implicit volume rendering regularization to bridge the gap between LiDAR and image data. 
% In contrast to these approaches, this study investigates an offboard setting, where aggregated LiDAR points, represented as binary occupancy, serve as input. Compared to LiDAR sweeps, binary occupancy data provides denser geometric details, including occluded regions, making it particularly advantageous in offboard scenarios.
In contrast to these approaches, our proposed method utilizes binary occupancy data instead of LiDAR point clouds, offering a stronger predictive performance.

\noindent
% {\bf Pretraining:} 
{\bf Learning-based auto-labeling.}
% In autonomous driving, it is necessary to prepare a large amount of ground truth data, such as 3D bounding box, LiDAR annotation, and 3D occupancy. 
% A best practice for this involves training a teacher machine learning model on annotated data and using the model's predictions to assign pseudo-labels to unannotated data, a method known as learning-based auto-labeling. 
% For instance, in building nuPlan dataset~\cite{karnchanachariLearningbasedPlanningNuPlan2024, caesarNuPlanClosedloopMLbased2022}, a large-scale teacher neural network was employed to generate the ground truth for 3D bounding boxes from LiDAR point clouds.
% However, to the best of our knowledge, no attempts have been made to add semantic annotations to unannotated binary occupancy data to create a 3D occupancy dataset in a learning-based auto-labeling setting.
Auto-labeling is a crucial strategy for creating large-scale GT data~\cite{karnchanachariLearningbasedPlanningNuPlan2024c, caesarNuPlanClosedloopMLbased2022} in autonomous driving systems. 
A common practice involves using offboard models, which are not subject to causality constraints and have minimal restrictions on inference speed. 
In the context of 3D object detection~\cite{qi2021offboard,ma2023detzero,fan2023once}, offboard models have been proposed and have demonstrated significantly superior performance compared to online models. However, to the best of our knowledge, no prior work has explored an offboard approach for semantic occupancy prediction. In this study, we propose a novel method that utilizes binary occupancy data in an offboard setting.

% \section{Methodology}
\section{Proposed Approach}
This section is structured as follows. First, the problem setting of this study is described in Sec.~\ref{sec:problem_setting}. Next, an overview of the proposed model is provided in Sec.~\ref{sec:model_overview}. Then, the utilization of the proposed model for binary occupancy pre-training (Sec.~\ref{sec:binary_occ_pretraining}) and learning-based auto-labeling (Sec.~\ref{sec:learning_based_auto_labeling}) is explained.

% \subsection{Problem Setup}
\subsection{Problem Setting}
\label{sec:problem_setting}
Our objective is to predict the semantic occupancy surrounding the ego-vehicle using $N$ surround-view RGB images with given camera poses and intrinsic parameters. The semantic occupancy of a scene $i$ is represented by a 3D grid $Y_i^s\in \{c_0, c_1, \dots, c_K\}^{H\times W\times Z}$, which is defined in the ego-coordinate system. Each voxel within this grid is assigned either a ``free'' label (denoted as $c_0$) or a semantic class label from the set $\{c_1, \dots, c_K\}$. In contrast, the binary occupancy $Y_i^b$ consists of ``free'' and ``occupied'' voxels, formally expressed as $Y_i^b\in\{0, 1\}^{H\times W\times Z}$.

In this study, we consider a situation where the available GT data for binary occupancy is significantly more abundant than that for semantic occupancy. This situation is commonly encountered in practical applications, as the generation of GT binary occupancy data does not necessitate semantic segmentation of LiDAR points. 
% Additionally, we assume that both the intrinsic and extrinsic parameters of the cameras are known for each frame.

\subsection{Model Overview}
\label{sec:model_overview}
% Figure \ref{fig:fig_concept} illustrates the core concept of this study. 
Figure~\ref{fig:fig_concept} illustrates the overview of the proposed method. 
We decompose the semantic occupancy prediction task into two subtasks: 1) binary occupancy prediction and 2) semantic occupancy prediction using the predicted binary occupancy. 
% Given that binary occupancy data is available in substantially larger quantities than semantic occupancy data, we can leverage this large-scale dataset to enhance component (1). 
Given that binary occupancy data are available in substantially larger quantities than semantic occupancy data, we can leverage this large-scale dataset to improve the binary occupancy prediction task. 
Our strategy employs such precise binary occupancy predictions as input for the subsequent semantic occupancy prediction module which is specifically designed to enhance the effectiveness of the binary occupancy data. 

As illustrated in Fig.~\ref{fig:fig_concept}, multi-view image features are first extracted using an image encoder. These features are then transformed into a 3D voxel feature representation, denoted as $B\in\mathbb{R}^{H^c\times W^c\times Z^c\times C}$, via an image-to-3D transformation module (explained in detail later). The model subsequently employs a binary decoder using the 3D voxel feature as input to identify occupied regions.
Here, $H^c$, $W^c$, and $Z^c$ represent the spatial dimension of the 3D representation, and $C$ denotes the feature dimension.
More specifically, $B$ is upsampled to $B'\in\mathbb{R}^{H^b\times W^b\times Z^b\times C'}$ and then used to predict the binary state of each voxel in $B'$. This process yields a set of feature vectors, denoted as $\mathcal{B}=\{b_i\}_{i=1}^{M}$, where $M$ represents the number of occupied voxels predicted in $B'$. Notably, we adopt a 3D compact representation~\cite{maCOTRCompactOccupancy2024a} for $B$, as the BEV representation lacks the $Z$ dimension, making it unsuitable for identifying feature vectors corresponding to occupied voxels and effectively incorporating them into subsequent modules. Please refer to Appendix~\ref{sec:implemetation_details_app} for more details.

Following this step, a transformer decoder, referred to as sparse transformer, is applied to $\mathcal{B}$ to refine its semantic representation. Consistent with \cite{li2024bevformer}, the sparse transformer composes deformable self/cross-attention and MLP. Since cross-attention and MLP computations are defined for each element of $\mathcal{B}$, the original transformer decoder \cite{li2024bevformer} can be directly applied to $\mathcal{B}$. However, deformable self-attention cannot be directly utilized for sparse representations such as $\mathcal{B}$. Therefore, we instead apply deformable cross-attention \cite{zhu2021deformable}, where each element of $\mathcal{B}$ serves as the value, while the dense representation $B'$ is utilized as the query and key.
Finally, the enhanced $\mathcal{B}$ is concatenated to the dense representation $B'$ and upsampled to $B''\in\mathbb{R}^{H\times W\times Z\times C''}$ to predict semantic occupancy.

\begin{figure}[hbt]
\centering
\includegraphics[width=0.48\textwidth]{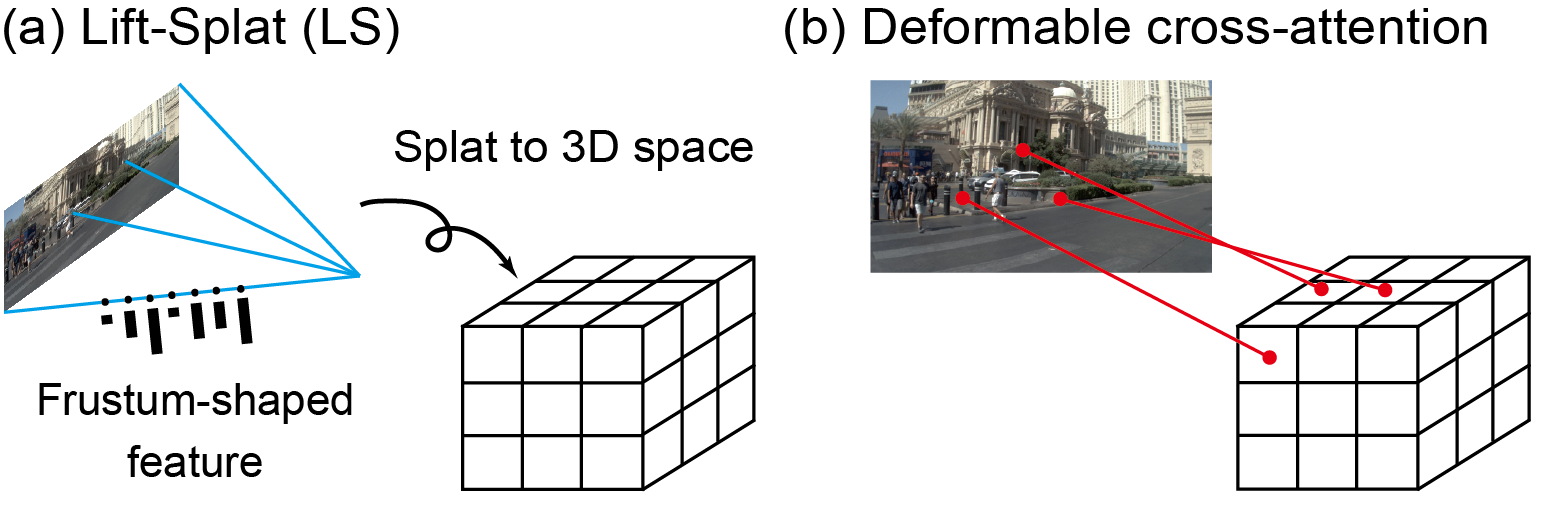}
\caption{Schematic of Image-to-3D transformation modules. (a) Lift-Splat (LS) and (b) deformable cross-attention modules. The voxel grids represent the 3D voxel representation. }
\label{fig:fig_models}
\end{figure}

Most vision-based 3D perception methods rely on one or both of two image-to-3D transformation modules: Lift-Splat (LS)~\cite{philionLiftSplatShoot2020b} and deformable cross-attention~\cite{zhu2021deformable} (Fig.~\ref{fig:fig_models}). In this study, we integrate the proposed pre-training module with both transformation mechanisms to ensure broad applicability. BEVFormer~\cite{li2024bevformer} employs deformable cross-attention, along with deformable self-attention and MLP layers. We refer to this configuration as the BEVFormer-based method, while models using the Lift-Splat mechanism are referred to as the LS-based method.

\subsection{Binary Occupancy Pre-training}
\label{sec:binary_occ_pretraining}

To effectively utilize large-scale binary occupancy data, we adopt a two-stage training strategy. 
In the first stage, training is conducted up to the binary decoder using only binary occupancy data. 
% In the second stage, we fine-tune the entire model undergoes fine-tuning with both binary and semantic occupancy data. 
In the second stage, we fine-tune the entire model with both binary and semantic occupancy data. 
% The proposed model is designed such that the binary occupancy prediction module is placed in the middle of the overall process. 
% This architectural choice provides two key advantages during training. 
Our model architecture where the binary decoder is placed before the semantic prediction module has two key advantages in training. 
First, it enables efficient pre-training without negatively impacting the subsequent modules responsible for semantic occupancy prediction. Second, during the second stage of training, the same binary occupancy data utilized in pre-training can be leveraged to preserve the performance of the binary decoder. For the first stage, we employ a binary cross-entropy loss $L_{\rm{binary}}$ to optimize the binary occupancy prediction using $B'$. In the second stage, a focal loss $L_{\rm{semantic}}$ is utilized for semantic occupancy prediction with $B''$, in conjuction with $L_{\rm{binary}}$.

There may be other strategies which aim to utilize large-scale binary occupancy data. 
For instance, by separating the classification heads for binary and semantic occupancy prediction, binary occupancy data could be directly employed in both pre-training and fine-tuning. 
However, our experimental results indicate that this approach is suboptimal (see Sec.~\ref{sec:exp_binary_occupancy_pretraining} for further details).

\begin{comment}
To investigate the effectiveness of our strategy, we compared it with another possible strategy utilizing large-scale binary occupancy data where by separating the classification heads for binary and semantic occupancy prediction, binary occupancy data could be directly employed in both pre-training and fine-tuning. 
\end{comment}

\begin{comment}
A more straightforward architectural approach could be considered for utilizing large-scale binary occupancy data. 
For instance, by separating the classification heads for binary and semantic occupancy prediction, binary occupancy data could be directly employed in both pre-training and fine-tuning. 

However, our experimental results indicate that this approach is suboptimal (see Section 4 for further details).
\end{comment}

\subsection{Learning-based Auto-labeling}
\label{sec:learning_based_auto_labeling}
% In the auto-labeling of ground truth of 3D semantic occupancy, real-time prediction is not required.
Auto-labeling GT of 3D semantic occupancy is critical in practice, as it enables the preparation of large-scale datasets that exceed the feasibility of manual annotation. 
% In contrast to semantic occupancy prediction models employed in self-driving cars, real-time prediction is not required in auto-labeling. 
Unlike semantic occupancy prediction in self-driving application, where online prediction is required, auto-labeling does not impose such constraints.
In the context of 3D object detection, multi-modal models—commonly integrating image and LiDAR inputs—are frequently utilized as auto-labeling machines. 
% Similarly, in the domain of semantic occupancy prediction, several studies have investigated the application of multi-modal approaches. 
In contrast, while several studies have explored multi-modal approaches for semantic occupancy prediction, existing research has primarily focused on online settings.
For the specific purpose of auto-labeling of semantic occupancy prediction, an offboard setting must be considered to maximize the prediction performance.
Since causality constraints do not apply in this context, binary occupancy data can serve as an alternative to LiDAR sweeps.
Given that binary occupancy data offers denser geometric details, particularly at greater distances from the ego-vehicle, it is anticipated to be more suitable for auto-labeling tasks.

% \noindent
% {\bf Architecture.}
The proposed architecture for the auto-labeling setting is fundamentally the same as the online model described in Sec.~\ref{sec:binary_occ_pretraining}. 
The primary distinction lies in the use of GT binary occupancy as input. 
Unlike the online model, the offboard model does not predict binary occupancy; instead, it leverages GT binary occupancy to identify occupied regions.
The corresponding feature vectors for these occupied regions are then fed into the subsequent sparse transformer, similar to the online model.
For training, only the focal loss $L_{\rm{semantic}}$ is employed.

\section{Experiments}

\subsection{Binary Occupancy Pre-training}
\label{sec:exp_binary_occupancy_pretraining}

In this section, we experimentally validate the effectiveness of the proposed binary occupancy pre-training. To assess its impact, we evaluate the performance of semantic occupancy prediction using the pretrained model as the initial weights.

\noindent
{\bf Dataset.}
We leveraged two occupancy datasets: OpenScene~\cite{yang2023vidar,openscene2023,sima2023_occnet} and Occ3D~\cite{tianOcc3DLargeScale3D2023a}. 
% OpenScene is a subset of nuPlan dataset
OpenScene is the large-scale occupancy dataset generated from nuPlan~\cite{karnchanachariLearningbasedPlanningNuPlan2024c, caesarNuPlanClosedloopMLbased2022}. 
We randomly selected 8000 and 150 scenes for the training and validation datasets, respectively. 
Occ3D is an occupancy dataset derived from nuScenes~\cite{caesar2020nuscenes}, comprising 700 training scenes and 150 validation scenes.
Both datasets provide GT with a volume size of $200\times 200\times 16$.

\noindent
{\bf Implementation Details.}
The AdamW optimizer was utilized with a learning rate $2\times 10^{-4}$ and a weight decay of $1\times 10^{-2}$, and the same augmentation strategy as BEVDet was applied~\cite{huangBEVDetHighperformanceMulticamera2022}. 
Both the pre-training and fine-tuning procedures in this study were conducted over 24 epochs using a batch size of 8. 
For BEVFormer-based models, an image resolution of $450\times 800$ was utilized. 
To ensure a fair comparison, three dense transformer layers were applied in the original BEVFormer baselines, whereas the proposed method incorporated two dense transformer layers alongside one sparse transformer layer. 
For LS-based models, an image resolution of $256\times 704$ was utilized. 
The proposed method integrates one sparse transformer layer following the LS module, while the LS baselines utilize one dense transformer layer to maintain comparability. 
Further hyperparameter details can be found in Appendix~\ref{sec:implemetation_details_app}.

\begin{figure}[t!]
\centering
\includegraphics[width=0.48\textwidth]{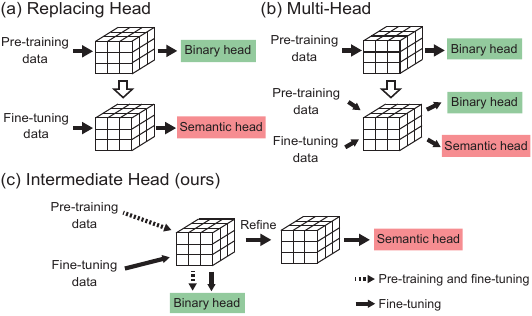}
\caption{Schematic of baseline strategies. The voxel grids represent the 3D voxel representation. }
\label{fig:fig_heads}
\end{figure}

\noindent
{\bf Baselines.}
During pre-training, all model parameters are updated in the baseline models. For fine-tuning, two baseline approaches are considered, as illustrated in Fig.~\ref{fig:fig_heads}. The first, {\it Replacing Head}, replaces the binary occupancy head with a semantic occupancy head, using only semantic occupancy data (corresponding to UniScene~\cite{min2024multi}). The second, {\it Multi-Head}, adds a new semantic occupancy head while retaining the binary occupancy head, preserving some of the pre-trained binary occupancy capabilities. In comparison, our method, {\it Intermediate Head}, places the binary occupancy head in intermediate stages of the process. It offers two optimization options for binary occupancy loss: (1) training exclusively on binary occupancy data derived from the semantic occupancy labels (denoted as B$^*$ in Table~\ref{table:comparison_with_baselines}), and (2) incorporating both pre-training binary occupancy data and binary occupancy data from the semantic labels (denoted as B). For the second option, a random batch of pre-training samples is selected to train the model up to the binary decoder.

% During pre-training, all model parameters are updated in the baseline configurations. For fine-tuning, two baseline approaches are considered.
% % , as illustrated in Fig.~\ref{fig:fig_heads}. 
% The first approach, referred to as {\it Replacing Head}, replaces the binary occupancy head with a semantic occupancy head, utilizing only semantic occupancy data. Note that this approach corresponds to UniScene~\cite{min2024multi}. The second approach, {\it Multi-Head}, adds a new semantic occupancy head while retaining the pretrained binary occupancy head. This strategy helps preserve, to some extent, the binary occupancy prediction capability acquired during pre-training. In comparison to these baselines, our method is referred to as {\it Intermediate Head}, as the binary occupancy head is positioned within the intermediate stages of the overall process. Additionally, the Intermediate Head strategy provides two optimization options for the binary occupancy loss. The first option involves training the model exclusively on binary occupancy data derived from the corresponding semantic occupancy data. The second option extends this approach by incorporating both the binary occupancy data used during pre-training and the binary occupancy data derived from the semantic occupancy labels. For the second option, a batch of samples is randomly selected from the pre-training dataset and used to train the model up to the binary decoder.

\begin{table*}[t!]
  \centering
  \resizebox{1.0\textwidth}{!}{
  \begin{tabular}{c|ccc|cc|cccccccc}
    \hline
    Method &  \begin{tabular}{c} Pre-training \\ Data \end{tabular} &  \begin{tabular}{c} Fine-tuning \\ Strategy \end{tabular} &  \begin{tabular}{c} Fine-tuning \\ Data \end{tabular}  & IoU & mIoU & \rotatebox[origin=c]{90}{ Vechicle } & \rotatebox[origin=c]{90}{ C. Zone Sign } & \rotatebox[origin=c]{90}{ Bicycle } & \rotatebox[origin=c]{90}{ Generic Obj. } & \rotatebox[origin=c]{90}{ Pedestrian } & \rotatebox[origin=c]{90}{ Traffic Cone } & \rotatebox[origin=c]{90}{ Barrier } & \rotatebox[origin=c]{90}{ Background }  \\
    \hline
    \hline
   BEVFormer & B & Replacing Head & S & 39.03 & 12.77 & 29.90 & 0.00 & 0.00 & 17.17 & 13.40 & 2.35 & 1.32 & 38.01    \\
   BEVFormer & B & Multi-Head & S+B* & 38.91 & 11.75 & 28.67 & 0.00 & 0.00 & 16.16 & 11.34 & 0.00 & 0.00 & 37.83   \\
     Sparse-BEVFormer (Ours)  & B & Intermediate Head & S+B*  & 40.09 & 15.12 & 31.13 & 0.00 & 6.27 & 18.27 & 15.63 & \textbf{3.72} & \textbf{6.83} & 39.09   \\
     Sparse-BEVFormer (Ours)  & B & Intermediate Head & S+B & \textbf{41.03} & \textbf{15.25} & \textbf{31.41} & 0.00 & \textbf{6.63} & \textbf{18.93} & \textbf{16.07} & 3.48 & 5.45 & \textbf{40.07}  \\
    \hline
   LS & B & Replacing Head & S & 43.08 & 18.35 & 36.35 & 0.01 & \textbf{8.66} & 21.73 & 19.46 & \textbf{10.22} & 8.27 & 42.07   \\
   LS & B & Multi-Head & S+B* & 43.12 & 16.00 & 36.02 & 0.00 & 1.04 & 20.74 & 18.74 & 7.5 & 1.95 & 42.03  \\
   Sparse-LS (Ours) &  B & Intermediate Head & S+B*  & 43.06 & 18.82 & 36.29 & 0.00 & 8.15 & 22.11 & 19.30 & 10.11 & 12.50 & 42.07 \\
   Sparse-LS (Ours) &  B & Intermediate Head & S+B & \textbf{43.92} & \textbf{19.20} &\textbf{36.88}&\textbf{0.23}&8.00&\textbf{22.92}&\textbf{19.49}&10.20&\textbf{12.94}&\textbf{42.94}  \\
    \hline
  \end{tabular}
  }
\caption{Quantitative comparison with baseline methods on OpenScene dataset. B and S represent binary occupancy data and semantic occupancy data, respectively. B$^*$ denotes binary occupancy data derived only from the semantic occupancy data used in the fine-tuning. 8,000 scenes from OpenScene were used for binary occupancy pre-training, while 500 scenes were utilized for semantic occupancy fine-tuning. All models were implemented based on COTR~\cite{maCOTRCompactOccupancy2024a} for these experiments.}
\label{table:comparison_with_baselines}
\end{table*}

\begin{figure}[t!]
\centering
\includegraphics[width=0.48\textwidth]{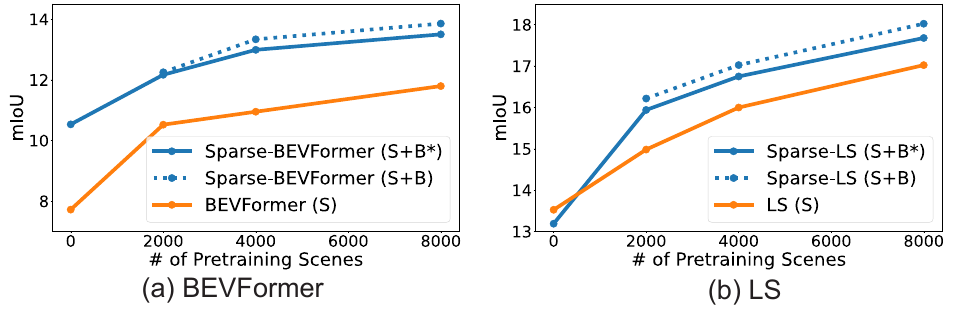}
\caption{Performance comparison of (a) BEVFormer-based and (b) LS-based methods on OpenScene.}
\label{fig:os_results}
\end{figure}

\subsubsection{Comparison with Baselines}
\label{sec:comparison_with_baselines}

Table~\ref{table:comparison_with_baselines} presents the quantitative comparison with the baseline methods.
The proposed fine-tuning strategy, Intermediate Head, achieved the best overall performance.
Among the methods, the Replacing Head approach is the most straightforward but yielded suboptimal results.
While the Multi-Head strategy also incorporates binary occupancy data during fine-tuning—similar to the Intermediate Head—it was outperformed by both other methods.
Additionally, we found that the Intermediate Head strategy further benefits from utilizing the binary occupancy data used during pre-training, which led to even greater performance improvements (denoted as S$+$B). 
We attribute this to the mitigation of performance degradation in binary occupancy prediction during fine-tuning under this setup, as further detailed in Appendix~\ref{sec:app_binary}. 
While the BEVFormer-based and LS-based methods exhibit some quantitative differences, the general trends observed across all strategies are consistent.

%Table~\ref{table:comparison_with_baselines} presents the quantitative comparison with the baseline methods. The Replacing Head strategy is the most straightforward; however, its performance was suboptimal. In contrast, the proposed fine-tuning strategy, Intermediate Head, achieved superior performance. Although the Multi-Head strategy also incorporates binary occupancy data during fine-tuning, its performance was inferior to the other methods. Furthermore, the Intermediate Head strategy benefits from leveraging the binary occupancy data used during pre-training, resulting in even greater performance improvements. In fact, the performance degradation of binary occupancy prediction during fine-tuning was mitigated, as demonstrated in Appendix \ref{sec:app_binary}.

Figure~\ref{fig:os_results} demonstrates that increasing the number of pre-training scenes consistently enhances model performance.
As shown in Table~\ref{table:comparison_with_baselines}, the proposed method outperforms the corresponding baselines for both BEVFormer- and LS-based models.
For LS-based methods, the baseline and the proposed approach perform similarly when pre-training is absent; however, with more pre-training scenes, the proposed method achieves superior performance.
This result highlights that our proposed architecture is particularly well-suited for leveraging pre-training strategies.
Given that the evaluation accuracy in Fig.~\ref{fig:os_results} has not yet saturated, even with 8,000 scenes, it is anticipated that performance will continue to improve with additional pre-training data. 
For the experiments shown in Fig.~\ref{fig:os_results}, 250 scenes were used for fine-tuning; additional results using 500 and 1000 fine-tuning scenes are provided in the Appendix~\ref{sec:app_several_fine_tune_scenes}.

\begin{figure}[t!]
\centering
\includegraphics[width=0.48\textwidth]{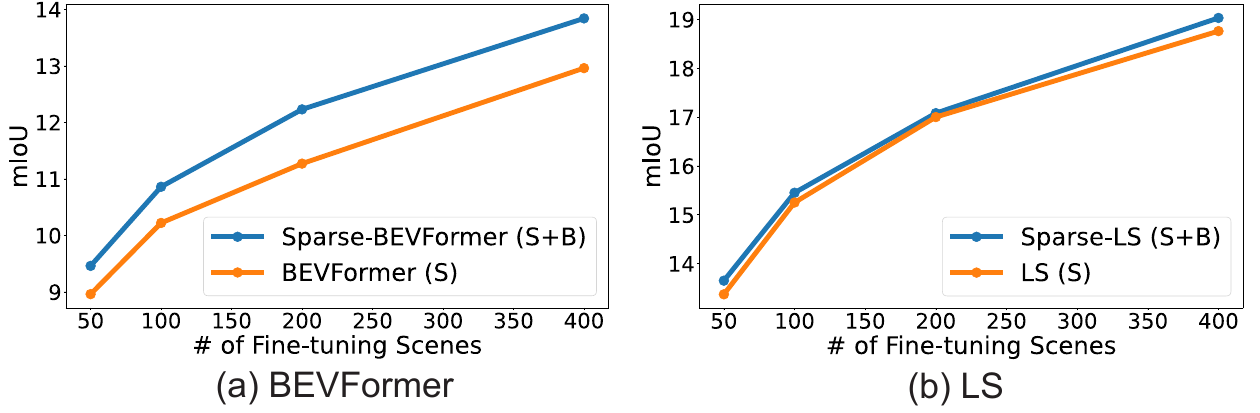}
\caption{Performance comparison of (a) BEVFormer-based and (b) LS-based methods on nuScenes.}
\label{fig:ns_results}
\end{figure}

\subsubsection{Comparison on nuScenes}
Figure~\ref{fig:ns_results} presents the results of experiments conducted on nuScenes. Unlike the OpenScene dataset, nuScenes does not contain a sufficient amount of training data to replicate the experimental conditions described in Sec.~\ref{sec:comparison_with_baselines}. Therefore, the objective of this section is to confirm the consistency of the obtained results with those reported in Sec.~\ref{sec:comparison_with_baselines}. Specifically, we fixed the number of pre-training scenes at 700 and varied the number of fine-tuning scenes. The proposed method demonstrated superior performance in BEVFormer-based scenarios, while achieving competitive performance in LS-based scenarios. These results are consistent with those in Sec.~\ref{sec:comparison_with_baselines} when comparing experiments with a similar number of pre-training scenes. More specifically, the crossover point between the Sparse-LS and LS results in Fig.~\ref{fig:os_results}(b) corresponds to experiments conducted with approximately 500 pre-training scenes and 250 fine-tuning scenes. This experimental condition is comparable to that of Fig.~\ref{fig:ns_results}(b). Therefore, as the number of pre-training scenes increases, the superiority of the proposed method is expected to become more pronounced.

\begin{figure}[t!]
\centering
\includegraphics[width=0.48\textwidth]{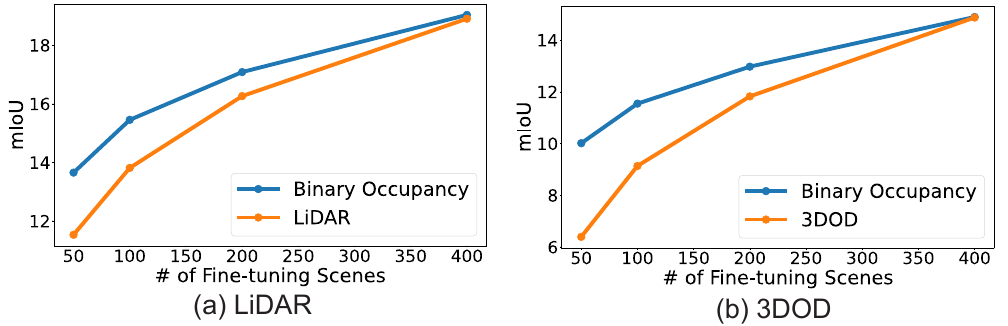}
\caption{Comparison with (a) LiDAR and (b) 3D object detection (3DOD) pre-training methods.}
\label{fig:ns_other_pretrain}
\end{figure}

% \begin{table}[t!]
%   \centering
%   \begin{minipage}[t]{0.45\textwidth}
%     \centering
%     \resizebox{0.99\textwidth}{!}{
%   \begin{tabular}{c|c|ccc}
%     \hline
%     Pretraining Data & Method & IoU &  mIoU & Imput Size   \\
%     \hline
%     \hline
%    ImageNet & Sparse-LSS  &  &  & $900\times 1600$   \\
%    3DOD &   &  &  & $900\times 1600$   \\
%    LiDAR &   &  &  &  $900\times 1600$   \\
%    \hline
%    Binary occupancy &  & &  &   \\
%     \hline
%   \end{tabular}
%   }
%   \subcaption{}
%   \end{minipage}
%   \\
% \begin{minipage}[t]{0.45\textwidth}
%   \centering
%   \resizebox{0.99\textwidth}{!}{
%   \begin{tabular}{c|cc|cc}
%     \hline
%     Pretraining Data & IoU &  mIoU & Backborn & Imput Size   \\
%     \hline
%     \hline
%    ImageNet &   &  &  & $900\times 1600$   \\
%    3DOD &   &  &  & $900\times 1600$   \\
%    LiDAR &   &  &  &  $900\times 1600$   \\
%    \hline
%    Binary occupancy &  & &  &   \\
%     \hline
%   \end{tabular}
%   }
%   \subcaption{}
%   \end{minipage}
% \caption{Comparison with (a) LiDAR and (b) 3DOD pretraining methods.}
% \label{table:comparison_with_other_pretraining}
% \end{table}

\subsubsection{Comparison with Other Pre-training Methods}
We compared the proposed binary occupancy pre-training approach with prevalent pre-training methods. Many semantic occupancy prediction methods utilize initial weights pre-trained on 3D object detection using the nuScenes dataset, as provided in \cite{wang2021fcos3d}. Additionally, pre-training using a depth loss based on LiDAR is widely adopted, as it does not require manual annotation. 
For this comparison, we pre-trained on 700 scenes from nuScenes and varied the number of scenes for fine-tuning. As Figure~\ref{fig:ns_other_pretrain} demonstrates, our proposed binary occupancy pre-training outperforms existing methods, especially when the amount of fine-tuning data is small. Although the performance gap diminishes as the number of fine-tuning scenes approaches that of the pre-training scenes, this is a general tendency in deep learning model training~\cite{he2019rethinking,liu2022improved}.
% Figure~\ref{fig:ns_other_pretrain} presents the qualitative evaluation results, demonstrating that the proposed binary occupancy pre-training approach outperforms existing methods when the number of fine-tuning scenes is small. This performance gap shrinks as the number of fine-tuning scenes increases.
Since the pre-training in \cite{wang2021fcos3d} was conducted with an image resolution of $900\times 1600$ and a ResNet101 backbone, the pre-training and fine-tuning in Fig.~\ref{fig:ns_other_pretrain}(b) were performed under the same conditions using the BEVFormer-based model to ensure a fair comparison.

\subsubsection{Qualitative Evaluation}
\label{sec:qualitative_eval}

\begin{figure*}[hbt]
\centering
\includegraphics[width=0.99\textwidth]{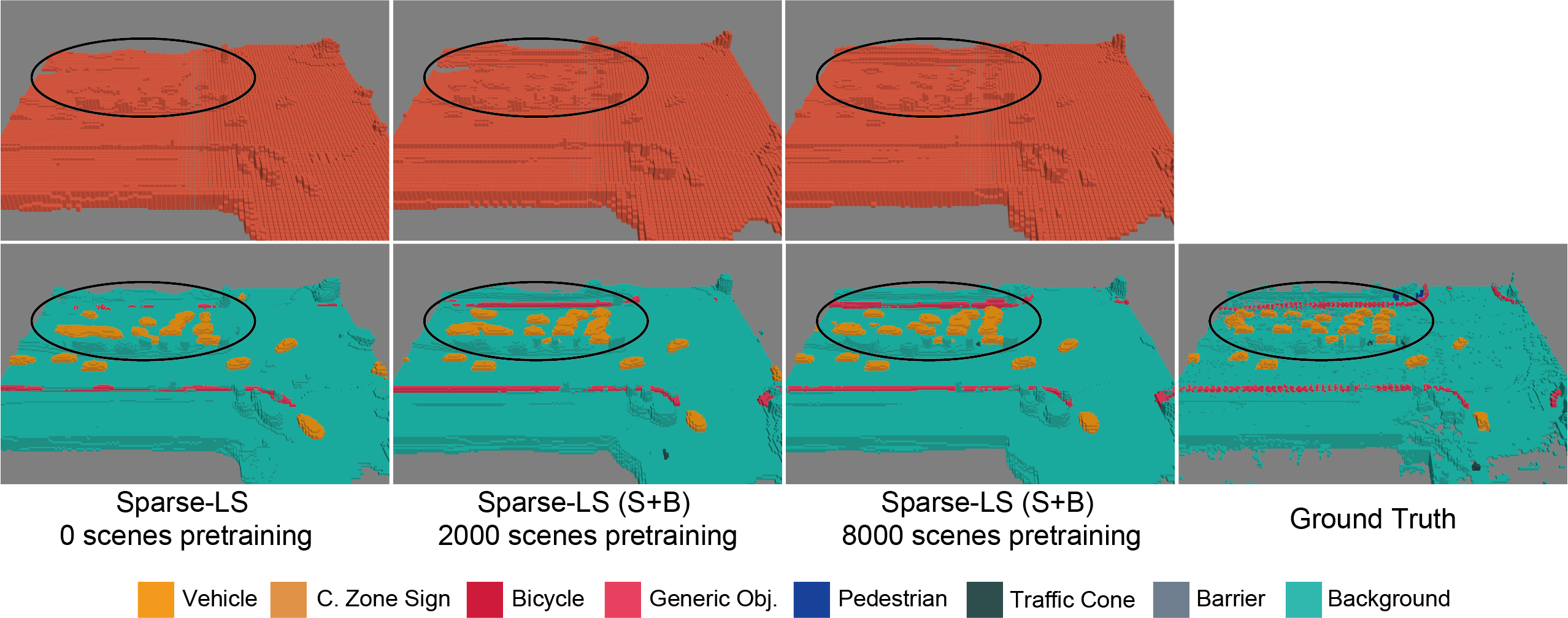}
\caption{Qualitative comparison across different numbers of pre-training scenes: 0, 2000, and 8000. The top row presents the outputs of the binary occupancy modules, while the bottom row shows the results of semantic occupancy prediction.}
\label{fig:qualitative_eval}
\end{figure*}

Figure~\ref{fig:qualitative_eval} presents a qualitative comparison across different numbers of pre-training scenes: 0, 2000, and 8000. The top row displays the outputs of the binary decoder, while the bottom row depicts the final outputs for semantic occupancy. Notably, as the number of pre-training scenes increases, the refinement of detailed structures becomes more pronounced. Additionally, the detailed structures in the outputs of the binary decoder also exhibit refinement. This observation suggests that the enhancement of geometric structures learned by the binary decoder contributes to the improved geometric accuracy of semantic occupancy.

\begin{table*}[t!]
  \centering
  \resizebox{0.99\textwidth}{!}{
  \begin{tabular}{c|c|cc|c|cccccccccccccccc}
    \hline
    Method & Modality &  IoU & mIoU & Offboard & \rotatebox[origin=c]{90}{ barrier } & \rotatebox[origin=c]{90}{ bicycle } & \rotatebox[origin=c]{90}{ bus } & \rotatebox[origin=c]{90}{ car } & \rotatebox[origin=c]{90}{ const. veh. } & \rotatebox[origin=c]{90}{ motorcycle } & \rotatebox[origin=c]{90}{ pedestrian } & \rotatebox[origin=c]{90}{ traffic cone } & \rotatebox[origin=c]{90}{ trailer } & \rotatebox[origin=c]{90}{ truck } & \rotatebox[origin=c]{90}{ drive. suf. } & \rotatebox[origin=c]{90}{ other flat } & \rotatebox[origin=c]{90}{ sidewalk } & \rotatebox[origin=c]{90}{ terrain } & \rotatebox[origin=c]{90}{ manmade } & \rotatebox[origin=c]{90}{ vegetation }    \\
    \hline
    \hline
   MonoScene~\cite{cao2022monoscene} & C  & 17.1 & 7.2 &  & 7.3 & 4.3 & 9.6 & 7.1 & 6.2 & 3.5 & 5.9 & 4.7 & 5.6 & 4.9 & 15.6 & 6.8 & 7.9 & 7.6 & 10.5 & 7.9  \\
   TPVFormer~\cite{huang2023tri} & C  & 15.1 & 8.3 &  & 9.7 & 4.5 & 11.5 & 10.7 & 5.5 & 4.6 & 6.3 & 5.4 & 6.9 & 6.9 & 14.1 & 9.8 & 8.9 & 9.0 & 9.9 & 8.5  \\
   3DSketch~\cite{chen3DSketchawareSemantic2020c} & C\&D  & 25.3 & 11.0 &  & 12.3 & 5.2 & 10.3 & 12.1 & 7.1 & 4.9 & 5.5 & 6.9 & 8.4 & 7.4 & 21.9 & 15.4 & 13.6 & 12.1 & 12.1 & 21.2  \\
   AICNet~\cite{liAnisotropicConvolutionalNetworks2020b} & C\&D  & 23.2 & 10.9 &  & 11.8 & 4.5 & 12.1 & 12.7 & 6.0 & 3.9 & 6.4 & 6.3 & 8.4 & 7.8 & 24.2 & 13.4 & 13.0 & 11.9 & 11.5 & 20.5  \\
   LMSCNet~\cite{roldaoLMSCNetLightweightMultiscale2020b} & L  & 26.7 & 11.8 &  & 12.9 & 5.2 & 12.8 & 12.6 & 6.6 & 4.9 & 6.3 & 6.5 & 8.8 & 7.7 & 24.3 & 12.7 & 16.5 & 14.5 & 14.2 & 22.1   \\
   JS3C-Net~\cite{yanSparseSingleSweep2021} & L  & 29.6 & 12.7 &  & 14.5 & 4.4 & 13.5 & 12.0 & 7.8 & 4.4 & 7.3 & 6.9 & 9.2 & 9.2 & 27.4 & 15.8 & 15.9 & 16.4 & 14.0 & 24.8  \\
   M-CONet~\cite{wangOpenOccupancyLargeScale2023b} & C\&L  & 26.5 & 20.5 &  & 23.3 & 16.1 & 22.2 & 24.6 & 13.3 & 20.1 & 21.2 & 14.4 & 17.0 & 21.3 & 31.8 & 22.0 & 21.8 & 20.5 & 17.7 & 20.4   \\
   Co-Occ~\cite{pan2024co} & C\&L  & 30.6 & 21.9 &  & 26.5 & \textbf{16.8} & 22.3 & 27.0 & 10.1 & \textbf{20.9} & 20.7 & \textbf{14.5} & 16.4 & 21.6 & 36.9 & 23.5 & 25.5 & 23.7 & 20.5 & 23.5   \\
   % OccMamba~\cite{li2024occmamba} & C\&L & 33.7 & 25.1 & & 29.6 & 20.2 & 25.7 & 28.5 & 16.7 & 25.0 & 23.2 & 19.9 & 20.3 & 24.5 & 36.1 & 25.3 & 25.1 & 24.8 & 27.7 & 28.9\\
   \hline
   Sparse-LS (Ours) & C\&B & \textbf{65.4} & \textbf{33.7} & \checkmark & \textbf{34.1} & 10.8 & \textbf{36.4} & \textbf{40.3} & \textbf{20.2} & 16.2 & \textbf{28.7} & 12.9 & \textbf{26.9} & \textbf{32.5} & \textbf{69.2} & \textbf{37.1} & \textbf{42.7} & \textbf{43.4} & \textbf{40.7} & \textbf{46.8}  \\
    \hline
  \end{tabular}
  }
\caption{Quantitative comparison on OpenOccupancy. The C, D, L,  and B denote camera, depth, LiDAR, and binary occupancy, respectively. ``Offboard'' indicates a model operating in an offboard setting.}
\label{table:comparison_with_baselines_offboard}
\end{table*}

\subsection{Learning-based Auto-labeling}
\label{sec:learning_based_auto_labeling}
In this section, we evaluate the effectiveness of the proposed auto-labeling strategy. To the best of our knowledge, the proposed method is the first to consider an offboard setting for semantic occupancy prediction. Therefore, we compare the proposed binary occupancy-based auto-labeling method with existing online models, including multi-modal models.

\noindent
{\bf Dataset.}
We utilized the OpenOccupancy dataset~\cite{wangOpenOccupancyLargeScale2023b} to evaluate offboard prediction performance. 
OpenOccupancy provides 17 semantic occupancy labels for nuScenes with a volume size of $512 \times 512\times 40$. 
Unlike other nuScenes-based occupancy benchmarks, the OpenOccupancy benchmark includes LiDAR-based baselines, making it well-suited for the objectives of this study.

\noindent
{\bf Implementation Details.}
We applied the same optimizer settings as described in Sec.~\ref{sec:exp_binary_occupancy_pretraining}. Training was conducted for 64 epochs with a batch size of 8.
The LS-based method was selected here, as it is expected to achieve superior results. An image resolution of $256\times 704$ was utilized.

\subsubsection{Comparison with Existing Methods}

Table~\ref{table:comparison_with_baselines_offboard} presents a quantitative comparison with existing multi-modal methods. The results shows that the proposed method achieves superior performance compared to other methods.
However, its performance is lower for three classes corresponding to small objects (bicycle, motorcycle, and traffic cone)
compared to methods using both camera LiDAR inputs like M-CONet and Co-Occ. 
We attribute this to the larger voxel size required for occupancy representations due to computational constraints (e.g., in the OpenOccupancy dataset, 0.2 in for occupancy vs. 0.1 for typical LiDAR points), which can cause fine-grained structural details to be lost.
Future work may explore the use of higher image resolutions—other methods employ an image resolution of $900\times 1600$—or the integration of LiDAR inputs to address this limitation.

The method’s strong overall performance makes it well-suited for auto-labeling pipelines. Although an mIoU of 33.7 is insufficient for direct use, practical setups can leverage larger models, more extensive datasets, and test-time augmentation~\cite{moshkov2020test} to reach acceptable accuracy. 
Additional results showcasing the application of the proposed method in auto-labeling pipelines are provided in Appendix~\ref{sec:performance_eval_auto_labeling_app}.

\begin{table}[t!]
  \centering
  \resizebox{0.48\textwidth}{!}{
  \begin{tabular}{c|cccc}
  \hline
   Stage & Training Data & \begin{tabular}{c} Number of \\ Pre-training Scenes \end{tabular} & \begin{tabular}{c} Number of \\ Fine-tuning Scenes \end{tabular} & Training Time (GPU Hours) \\
    \hline
    \hline
    Pre-training & B  & 4000 & 0 & 544 \\
    Pre-training & B & 8000 & 0 & 1032 \\
    Fine-tuning  & S+B$^*$ & 0 & 250 & 32 \\
    Fine-tuning & S+B & 8000$\dag$ & 250 & 54 \\
    \hline
  \end{tabular}
  }
\caption{Computational cost of pre-training and fine-tuning Sparse-LS. B represents binary occupancy data and S represents semantic occupancy data. The notation B$^*$ indicates binary occupancy data that was created exclusively from the semantic data of the fine-tuning set. 8000$\dag$ indicates that 250 scenes were randomly selected from 8000 scenes for each epoch.}
\label{table:comp_cost}
\end{table}

\subsection{Computational Cost}
We report the computational time for the pre-training and fine-tuning stages in Table~\ref{table:comp_cost}. The pre-training on 8,000 scenes demanded over 10 days of processing on a system equipped with four A100 GPUs. In the fine-tuning stage, the use of the S+B dataset increased the required training time in comparison to the S+B$^*$ dataset, an increase attributable to the incorporation of the B data.

\section{Conclusion}

% In this study, we propose a training strategy and model architecture for the 3D semantic occupancy prediction task specifically designed to effectively utilize the binary occupancy dataset. 
% By separating the binary prediction module and semantic occupancy prediction module,
This study introduces a learning method and architecture for 3D semantic occupancy prediction, specifically designed to leverage cost-effective binary occupancy datasets. 
Our approach divides the problem into two sequential steps: binary occupancy prediction and semantic prediction. 
% In this study, we propose a learning method and architecture for a 3D semantic occupancy prediction task designed to efficiently utilize a binary occupancy dataset, which can be prepared without semantic annotation cost. 
% Specifically, we
% Our approach
% divides the 3D semantic occupancy prediction task into two steps: a binary occupancy prediction task and a semantic prediction task. 
The model is first pre-trained on large-scale binary GT, then fine-tuned using a combination of binary and semantic data. This pre-training and fine-tuning strategy achieves superior prediction accuracy compared to simpler alternatives like the Multi-Head and Replacing Head methods.
% We first conduct pre-training using a large-scale binary occupancy GT for the binary occupancy prediction module. 
% Subsequently, we perform fine-tuning with semantic data. 
% This strategy demonstrates superior prediction accuracy compared to other two-step learning methods, such as the Multi-Head and Replacing Head approaches. 
This method also functions as a learning-based auto-labeling machine using binary occupancy ground truth (GT), outperforming LiDAR or image-based models. Its higher accuracy likely stems from the richer information in binary occupancy data, including occlusion regions missed by LiDAR, and our model's effective use of this data.
% Additionally, this method offers the advantage of functioning as an learning-based auto-labeling machine using binary occupancy GT as input, achieving higher accuracy than models using LiDAR or image inputs. 
% This superior performance is likely due to the greater information of binary occupancy data, which includes occlusion regions typically missed by LiDAR, as well as our model architecture's ability to effectively leverage this information. 

% The performance of our model did not saturate with respect to the number of scenes used in the pre-training. 

{\small
\bibliographystyle{unsrt}
% \bibliography{egbib}
\bibliography{egbib_final_abbreviation}
}

\clearpage

\appendix

\twocolumn[
    \begin{center}
        {\Large \bfseries From Binary to Semantic: Utilizing Large-Scale Binary Occupancy Data \\ for 3D Semantic Occupancy Prediction} % Enlarged bold title
    \end{center}
    % \vspace{3pt}
    \begin{center}
        {\Large Supplementary Material} % Enlarged bold title
    \end{center}
    \vspace{10pt}
]

\section{Implementation Details}
\label{sec:implemetation_details_app}

\subsection{3D Representation}

As discussed in Sec.~\ref{sec:model_overview}, we employed a 3D compact representation rather than a BEV representation. The primary limitation of the BEV representation is the absence of the $Z$ dimension, while it retains full resolution along the $H$ and $W$ dimensions (typically $H=200$ and $W=200$). To perform binary occupancy prediction from the BEV representation, the channel dimension $C$ must be reshaped (e.g., $C=512$ is typically reshaped as $Z\times C'=16\times 32=512$). However, this transformation presents two challenges for our approach. 

First, the reduced channel dimension $C'=32$ is insufficient for the subsequent sparse transformer, potentially creating a bottleneck in the model. Second, the full resolution of the 3D space (typically $H=200$, $W=200$, and $Z=16$) is computationally prohibitive for real-time applications. Consequently, it is necessary to downsample the spatial resolution from $(H, W)=(200, 200)$ to $(H', W')=(100, 100)$, which may introduce an additional bottleneck in the model.

To mitigate these limitations, the proposed model adopts a 3D compact representation with $H^c=50$, $W^c=50$, $Z^c=16$, and $C=256$, thereby addressing the aforementioned concerns. The representation is then upsampled to a resolution of $H^b=100$, $W^b=100$, and $Z^c=16$ for binary occupancy prediction. For auto-labeling experiments, we employed higher-resolution setting with $H^c=H^b=256$, $W^c=W^b=256$, $Z^c=Z^b=40$, and $C=64$, as the OpenOccupancy dataset provides a larger volume size ($H=512$, $W=512$, $Z=40$), and the auto-labeling experiments consider only an offboard setting.

\subsection{Architecture Details}
For both LS-based and BEVFormer-based methods 
% (Fig.~\ref{fig:fig_models}),
we employed a ResNet50 backbone with a FPN as the neck. The standard transformer in BEVFormer utilizes spatially dense queries and is therefore referred to as a dense transformer in the main text. In contrast, the proposed method incorporates a sparse transformer, which utilizes feature vectors corresponding to occupied regions as queries, as explained in Sec.~\ref{sec:model_overview}. In both transformer architectures, we adopted the following hyperparameters: the number of attention heads is set to 8, the number of reference points for deformable attention is 4 in the self-attention module and 8 in the cross-attention module, and the embedding dimension is 256. For further implementation details, please refer to the provided code.

% \begin{figure}[hbt]
% \centering
% \includegraphics[width=0.48\textwidth]{Figs/fig_models/fig_models_ver2.png}
% \caption{Schematic of Image-to-3D transformation modules. (a) Lift-Splat (LS) and (b) deformable cross-attention modules. The voxel grids represent the 3D voxel representation. }
% \label{fig:fig_models}
% \end{figure}

\section{Evaluating Semantic Occupancy Prediction with Varying Numbers of Fine-tuning Scenes}
\label{sec:app_several_fine_tune_scenes}

\begin{figure*}[hbt]
\centering
\includegraphics[width=0.99\textwidth]{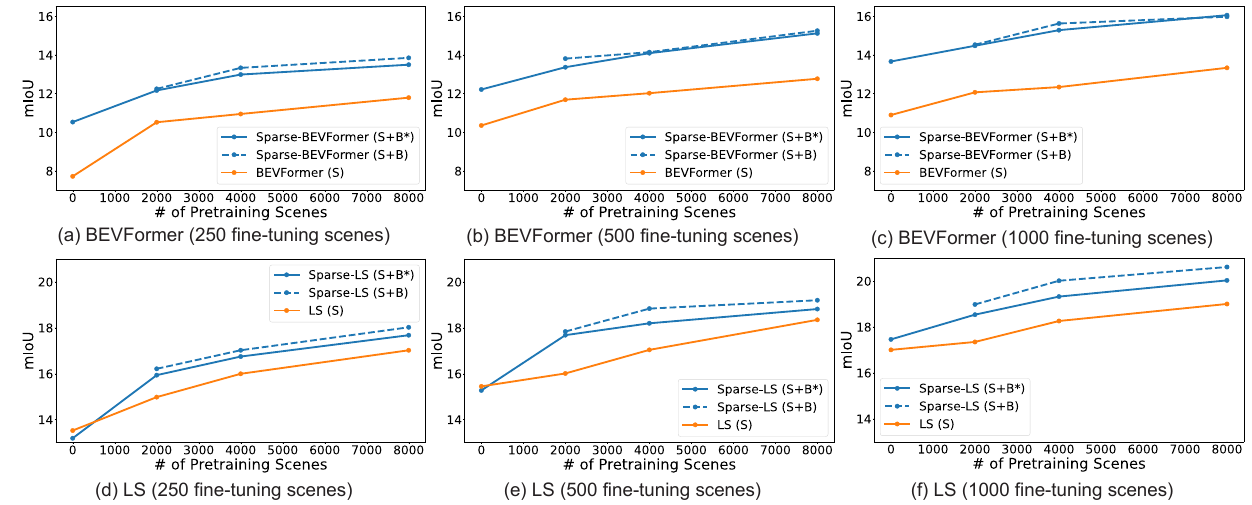}
\caption{These figures display the mIoU scores as the number of pre-training scenes varies. The top row shows the results of BEVFormer-based models, while the bottom row presents the results of LS-based models.}
\label{fig:os_results_app}
\end{figure*}

In Section~\ref{sec:comparison_with_baselines}, we presented experimental results using 250 fine-tuning scenes. Figure~\ref{fig:os_results_app} provides additional results with 500 and 1000 fine-tuning scenes. The proposed method consistently demonstrates a performance advantage across all fine-tuning set sizes.

\section{Evaluating Binary Occupancy Prediction During Pre-training}
\label{sec:app_binary}
\begin{figure*}[hbt]
\centering
\includegraphics[width=0.99\textwidth]{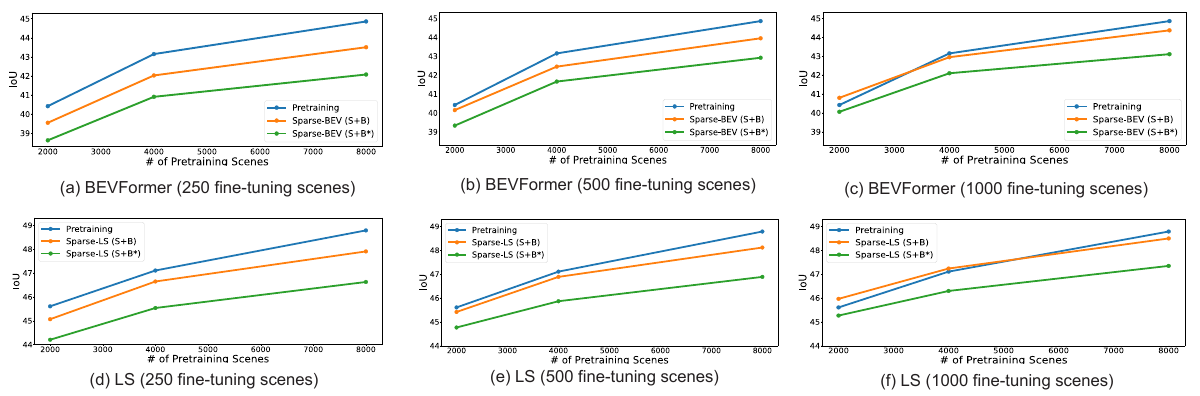}
\caption{These figures display the IoU scores of binary occupancy prediction module with the proposed method as a function of the number of pre-training scenes.}
\label{fig:fig_pt_results}
\end{figure*}

Figure~\ref{fig:fig_pt_results} provides the IoU scores of the binary occupancy prediction module within the proposed method as a function of the number of pre-training scenes. The IoU scores are highest upon the completion of pre-training but decrease during fine-tuning, as the number of fine-tuning scenes is substantially smaller than that of pre-training scenes. However, the proposed fine-tuning strategy effectively mitigates this performance degradation by incorporating the pre-training scenes during fine-tuning.

\section{Performance Evaluation in Auto-labeling Pipelines}
\label{sec:performance_eval_auto_labeling_app}
In Sec.~\ref{sec:learning_based_auto_labeling}, we evaluated the proposed auto-labeling model using the OpenOccupancy dataset. However, OpenOccupancy is not sufficiently large to evaluate auto-labeling pipelines. Therefore, in this section, we utilize the OpenScene dataset to generate a large-scale semantic occupancy dataset using the proposed pipeline and evaluate its effectiveness by training online methods on the generated dataset.

Table~\ref{table:os_offboard} presents a comparison of Sparse-LSS in an offboard setting—where GT binary occupancy is provided as input—versus an onboard setting. Consistent with the results on OpenOccupancy, the offboard model achieves significantly higher performance than its onboard counterparts.

Using the trained offboard model, we generated pseudo-labels. However, dense predictions such as semantic occupancy prediction require substantial storage capacity to save the generated labels. Ideally, storing logits for all voxels would maximize performance, but this is impractical due to storage constraints. Therefore, in our experiments, we compared two types of pseudo-labeling approaches. The first is a storage-efficient approach, where only label indices corresponding to the highest predicted probability are stored (denoted as ``Top1'' in Tab.~\ref{table:os_auto_labeling}). The second is a balanced approach, where the top two logits are stored (denoted as ``Top2'' in Tab.~\ref{table:os_auto_labeling}).

As a baseline, we trained an LS-based model on 500 scenes with GT semantic occupancy labels. From this baseline, we progressively increased the number of scenes with pseudo-labels. As shown in Tab.~\ref{table:os_auto_labeling}, both IoU and mIoU improve as the number of pseudo-labeled scenes increases. Additionally, as expected, the Top2 strategy outperforms the Top1 strategy, demonstrating the benefits of retaining more predictive information. Notably, when 7,500 pseudo-labeled scenes are used, the model's performance becomes competitive with that of a model trained on 2,000 scenes with GT labels, highlighting the effectiveness of pseudo-labeling in reducing reliance on manually annotated data.

\begin{table*}[t!]
  \centering
  \resizebox{0.98\textwidth}{!}{
  \begin{tabular}{c|c|cc|cccccccc}
  \hline
  Method & Offboard & IoU & mIoU & \rotatebox[origin=c]{90}{ Vechicle } & \rotatebox[origin=c]{90}{ C. Zone Sign } & \rotatebox[origin=c]{90}{ Bicycle } & \rotatebox[origin=c]{90}{ Generic Object } & \rotatebox[origin=c]{90}{ Pedestrian } & \rotatebox[origin=c]{90}{ Traffic Cone } & \rotatebox[origin=c]{90}{ Barrier } & \rotatebox[origin=c]{90}{ Background }  \\
    \hline
    \hline
  Sparse-LS & & 43.92 & 19.20 &36.88&0.23&8.00&22.92&19.49&10.20&12.94&42.94 \\
  Sparse-LS & \checkmark & \textbf{73.89} & \textbf{33.17} & 62.44 & 6.67 & 10.11 & 33.38 & 37.62 & 23.08 & 19.60 & 72.44  \\
    \hline
  \end{tabular}
  }
\caption{Performance comparison between offboard and onboard models on the OpenScene dataset. The onboard model was pretrained on 8000 scenes and fine-tuned on 500 scenes. The offboard model was trained on the same 500 scenes dataset.}
\label{table:os_offboard}
\end{table*}

\begin{table*}[t!]
  \centering
  \resizebox{0.92\textwidth}{!}{
  \begin{tabular}{c|cc|c|cc}
  \hline
  Method & \# of Scenes w/ GT & \# of Scenes w/ Pseudo-labels & Pseudo-labels & IoU & mIoU \\
    \hline
    \hline
  LS & 500 & 0 & - & 37.70 & 15.45\\
  LS & 2000 & 0 & - & 41.12 & 18.76 \\
    \hline
  LS & 500 & 1500 & Top1 & 37.62 & 17.10 \\
  LS & 500 & 1500 & Top2 & 44.62 & 17.40 \\
  LS & 500 & 3500 & Top1 & 38.24 & 17.55 \\
  LS & 500 & 3500 & Top2 & \textbf{46.09} & 18.11 \\
  LS & 500 & 7500 & Top1 & 38.72 & \textbf{18.22} \\
    \hline
  \end{tabular}
  }
\caption{Evaluation results of the auto-labeling pipelines using the proposed offboard model. Top1 denotes the auto-labeling approach in which only label indices corresponding to the highest predicted probability are stored, whereas Top2 represents the approach where the top two logits are stored.}
\label{table:os_auto_labeling}
\end{table*}

\section{Performance Comparison with Longer Epochs}
% All experiments in Sec.~\ref{sec:exp_binary_occupancy_pretraining} were conducted with 24 epochs.
% In this section, we present experimental results with an extended training duration of 64 epochs. Figure~\ref{fig:longer_epochs_supp} illustrates the mIoU values as a function of the number of pre-training scenes when 250 scenes were used for fine-tuning.

\begin{figure}[t!]
\centering
\includegraphics[width=0.46\textwidth]{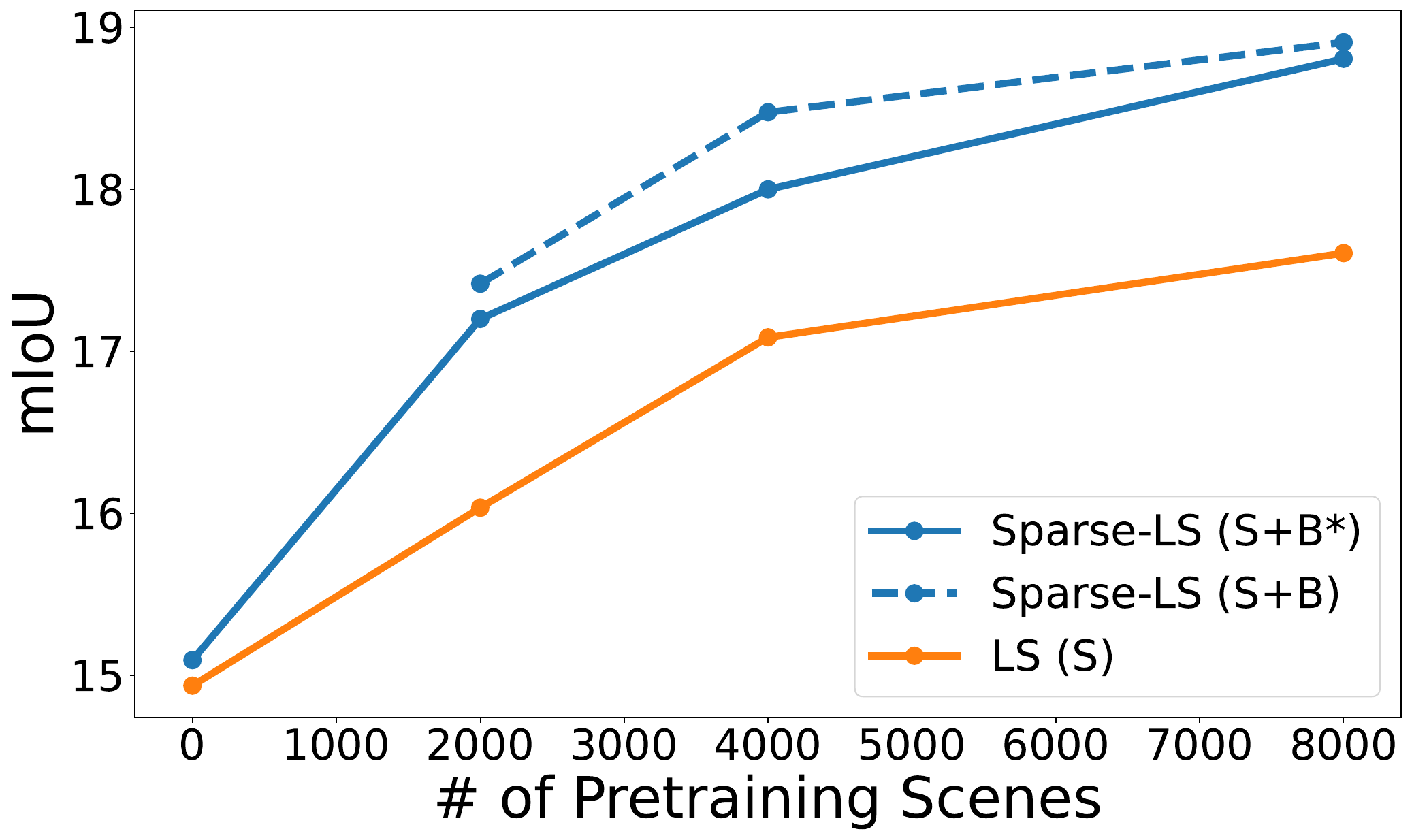}
\caption{mIoU scores at 64 epoch as a function of the number of pre-training scenes. 250 scenes are used for the fine-tuning.}
\label{fig:longer_epochs_supp}
\end{figure}

All experiments in Sec.~\ref{sec:exp_binary_occupancy_pretraining} were conducted with 24 epochs.
We here present experimental results with an extended fine-tuning duration of 64 epochs, during which the mIoU scores fully reach the plateau.
As illustrated in Figure~\ref{fig:longer_epochs_supp}, our strategy outperforms the baseline by achieving higher plateau mIoU scores.
% Figure~\ref{fig:longer_epochs_supp} shows our strategy is superior to the baseline for the plateau mIoU. 
% Figure~\ref{fig:longer_epochs_supp} illustrates the mIoU values as a function of the number of pre-training scenes when 250 scenes were used for fine-tuning. 
% Our strategy is superior to the baseline for the plateau mIoU. 

% All experiments in Sec.~\ref{sec:exp_binary_occupancy_pretraining} were conducted with 24 epochs. In this section, we present experimental results with an extended training duration of 64 epochs to further evaluate the superiority of the proposed approach. Figure~\ref{fig:longer_epochs_supp} illustrates the mIoU values as a function of the number of pre-training scenes when 250 scenes were used for fine-tuning. The results confirm that the same conclusion drawn from Fig.~\ref{fig:os_results} in the main text remains valid.

\section{Qualitative Evaluation}
\begin{figure*}[t!]
\centering
\includegraphics[width=0.89\textwidth]{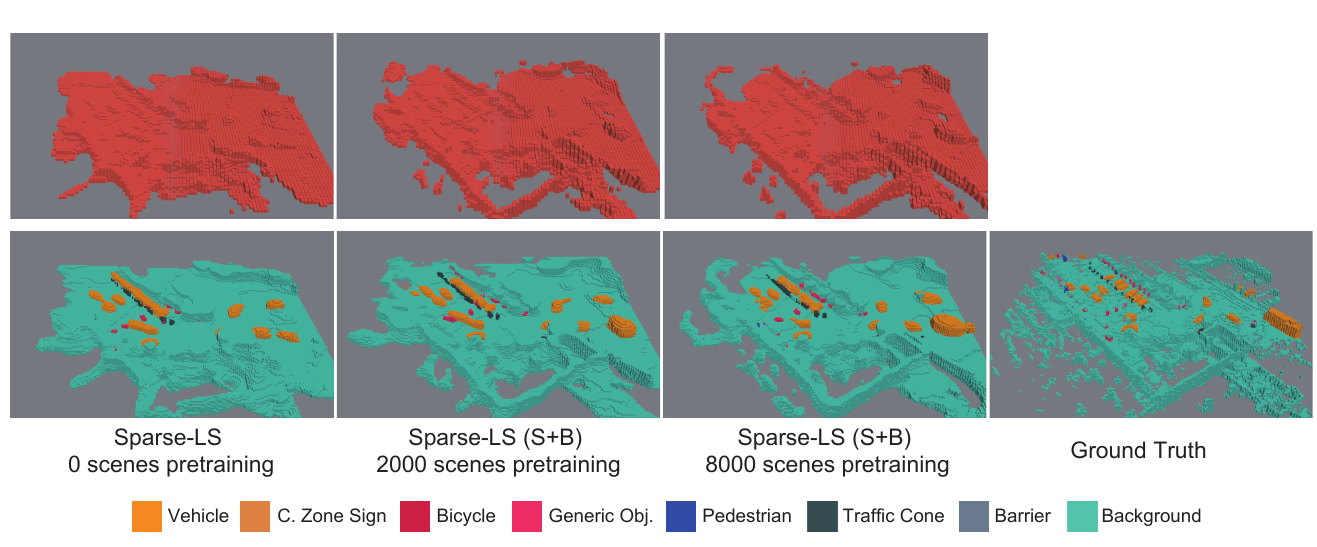}
\caption{Qualitative comparison across different numbers of pre-training scenes: 0, 2000, and 8000. The top row presents the outputs of the binary occupancy modules, while the bottom row shows the results of semantic occupancy prediction.}
\label{fig:qualitative_eval_supp1}
\end{figure*}

\begin{figure*}[t!]
\centering
\includegraphics[width=0.89\textwidth]{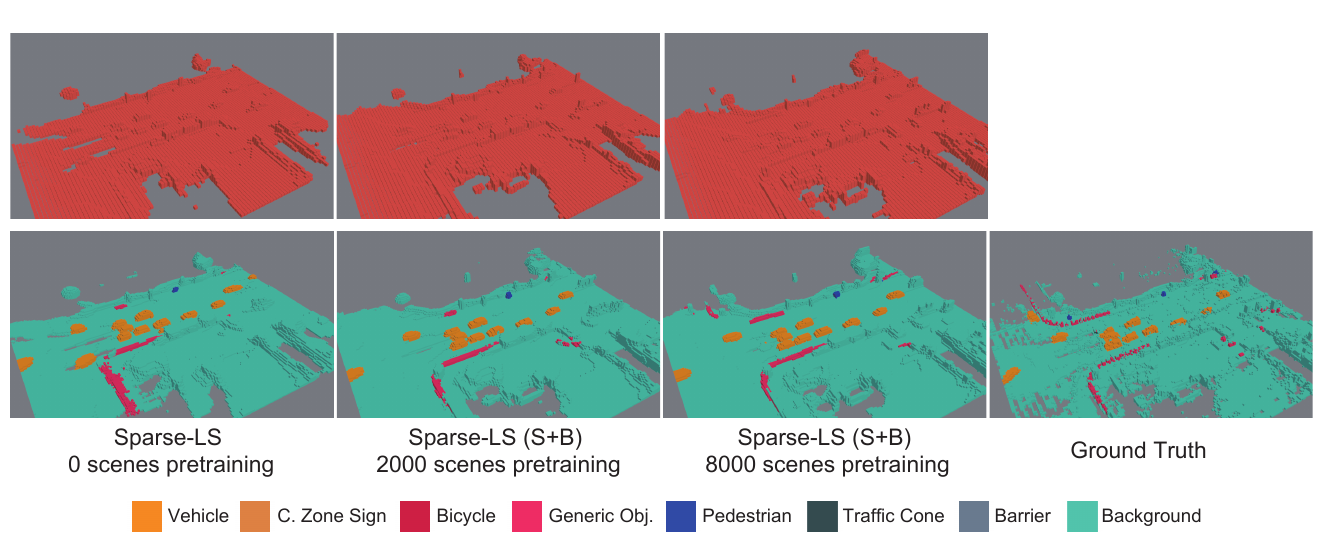}
\caption{Refer to the caption of Figure~\ref{fig:qualitative_eval_supp1} for a detailed description.}
\label{fig:qualitative_eval_supp2}
\end{figure*}

\begin{figure*}[t!]
\centering
\includegraphics[width=0.89\textwidth]{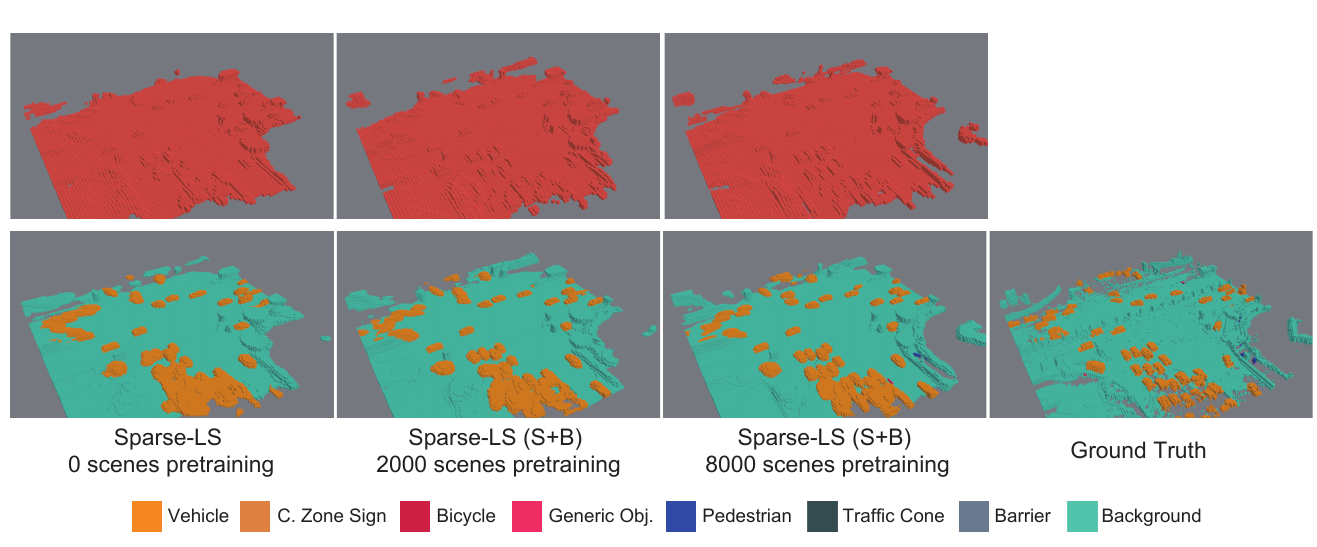}
\caption{Refer to the caption of Figure~\ref{fig:qualitative_eval_supp1} for a detailed description.}
\label{fig:qualitative_eval_supp3}
\end{figure*}

\begin{figure*}[t!]
\centering
\includegraphics[width=0.89\textwidth]{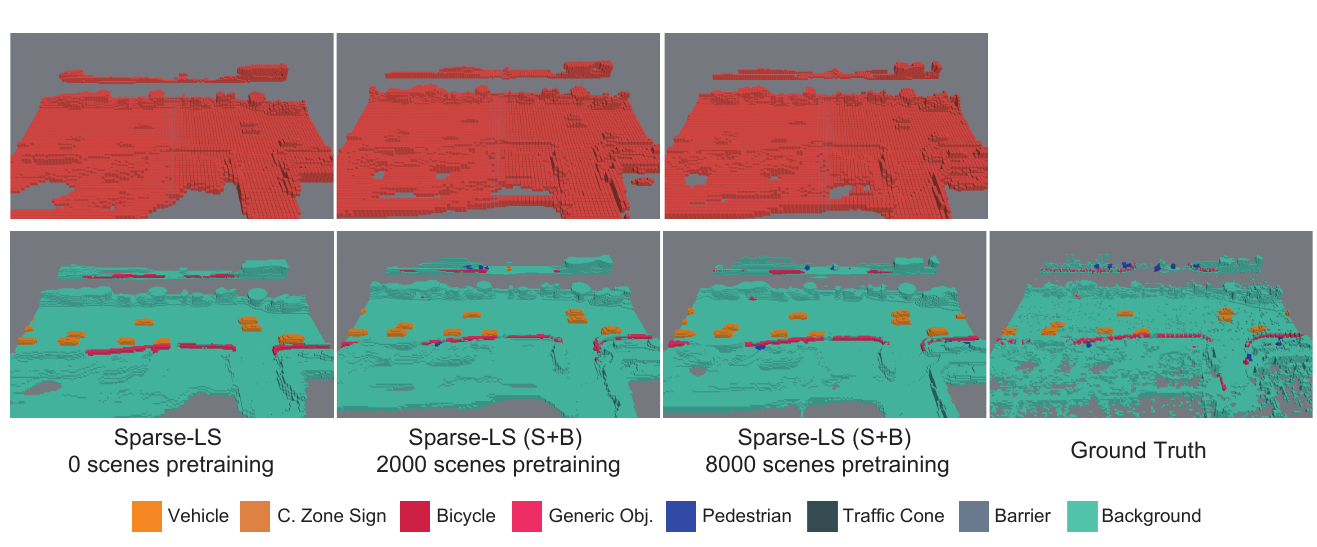}
\caption{Refer to the caption of Figure~\ref{fig:qualitative_eval_supp1} for a detailed description.}
\label{fig:qualitative_eval_supp4}
\end{figure*}

\begin{figure*}[t!]
\centering
\includegraphics[width=0.89\textwidth]{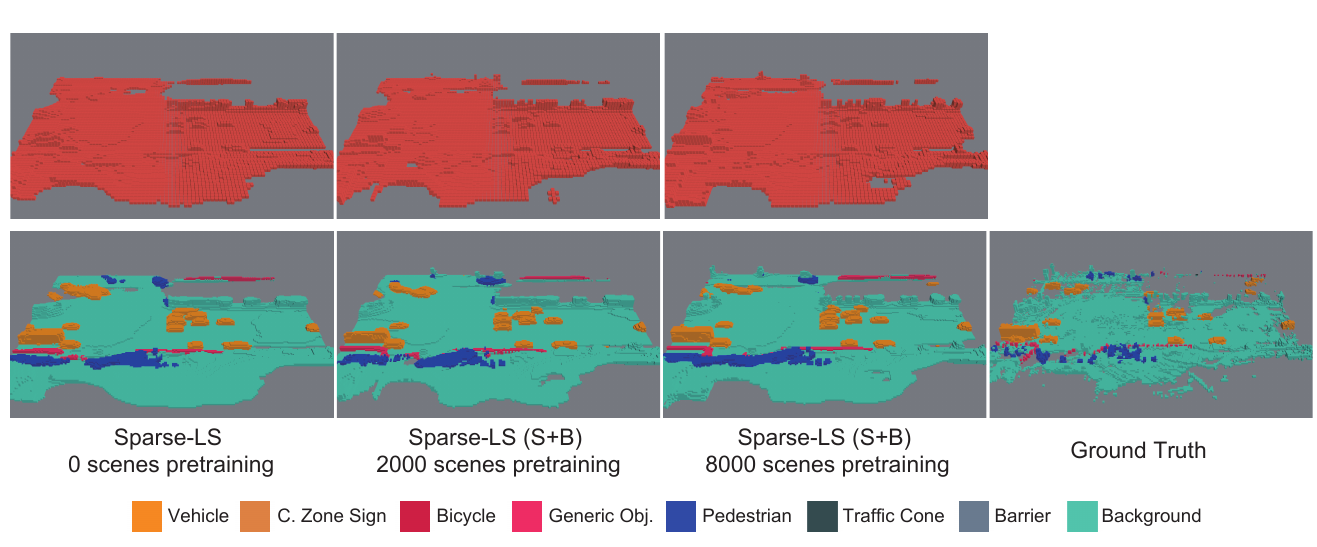}
\caption{Refer to the caption of Figure~\ref{fig:qualitative_eval_supp1} for a detailed description.}
\label{fig:qualitative_eval_supp5}
\end{figure*}

Figures~\ref{fig:qualitative_eval_supp1}-\ref{fig:qualitative_eval_supp5} present representative examples of the outputs generated by the proposed method, which are omitted from Sec.~\ref{sec:qualitative_eval} in the main text due to page limitations.

\end{document}